\documentclass[runningheads]{llncs}
\usepackage[T1]{fontenc}
\usepackage{amssymb}
\usepackage{amsmath}
\usepackage{subcaption}
\usepackage{multirow} 
\usepackage{graphicx}
\usepackage{hyperref}
\usepackage{cite}
\usepackage{amsfonts}
\begin{document}
\title{Complementary Roles of Radiomics and Foundation Representations in Renal Cell Carcinoma Classification: A Comparative Study of 2D and 3D CT Encodings}
\titlerunning{Radiomics and Foundation Representations for RCC Classification}
%
\author{
Yuan Liang\inst{1,3} \and
Sourav Bhattacharjee\inst{2} \and
Abraham Campbell\inst{3}
}
\authorrunning{Y. Liang et al.}
\institute{
Research Ireland Centre for Research Training in Machine Learning \and
School of Veterinary Medicine, University College Dublin, Dublin, Ireland \and
School of Computer Science, University College Dublin, Dublin, Ireland\\
\email{abey.campbell@ucd.ie}
}
\maketitle              
\begin{abstract}
Accurate preoperative subtype classification of renal cell carcinoma (RCC) from contrast-enhanced computed tomography remains clinically challenging. Radiomics provides structured tumour descriptors, whereas foundation representations offer transferable image features. However, it remains unclear whether radiomics still adds value beyond pretrained representations, and how 2D and 3D MedVAE encoders compare in this setting.

We compared handcrafted radiomics, 2D MedVAE, 3D MedVAE, and their fusion for binary clear-cell RCC versus non-clear-cell RCC classification on KiTS23 under a unified preprocessing pipeline. Concatenation, cross-attention, and gated fusion were evaluated as representative integration strategies, and radiomics feature importance was analysed to support decision-centric interpretability.

Fusion consistently improved discrimination over image-only MedVAE branches. The best overall performance was achieved by 3D gated fusion, with an AUC of 82.7\%, outperforming the best 2D fusion model (79.6\%), the radiomics baseline (74.4\%), and the single-modality MedVAE branches. Ablation analysis further showed clear gains of the full fusion model over both image-only and radiomics-only variants, indicating complementary contributions from radiomics and image representations.

These findings suggest that radiomics remains relevant for RCC CT classification in the presence of foundation representations, and that its integration with MedVAE is more effective in the 3D setting. More broadly, the study supports a complementary role for radiomics and foundation representations in clinically meaningful imaging decision support.

\end{abstract}
\section{Introduction}

Kidney cancer remains a substantial global health burden, with 434,840 new cases and 155,953 deaths reported worldwide in 2022~\cite{iarc2024kidneyfactsheet}. Accurate preoperative characterisation of renal masses, including discrimination between clear cell RCC (ccRCC) and non-clear cell RCC, is clinically important because histologic subtype is associated with prognosis and treatment decisions. Multiphasic contrast-enhanced CT is central to diagnosis and staging and is recommended by the European Association of Urology guideline for RCC~\cite{ljungberg2023eau}. However, CT interpretation remains limited by inter-reader variability and the difficulty of quantifying subtle intratumoral heterogeneity, motivating computational approaches that can provide reproducible and quantitative decision support.

Radiomics offers a principled way to quantify tumour phenotype from routine imaging. Early work established radiomics as high-throughput extraction of engineered intensity, shape, and texture descriptors and demonstrated associations with cancer phenotype and outcomes~\cite{aerts2014decoding}. Subsequent reviews outlined a standardised pipeline spanning image acquisition, segmentation, preprocessing, feature extraction, model development, and validation, while highlighting challenges such as protocol variability and overfitting~\cite{gillies2016radiomics,lambin2017radiomics}. Reproducibility has been further supported by the Image Biomarker Standardisation Initiative and tools such as PyRadiomics~\cite{zwanenburg2020image,van2017computational}. In RCC, CT radiomics has shown promise for distinguishing ccRCC from non-clear cell RCC, supporting its value as a structured and interpretable descriptor of renal tumour phenotype~\cite{wang2021radiomics}.

Deep learning has also become dominant across medical image analysis tasks~\cite{litjens2017survey}, and end-to-end models have been explored for differentiating renal tumour subtypes on multi-phase CT~\cite{uhm2021deep}. Yet RCC studies often involve limited labelled cohorts and heterogeneous acquisition protocols, making robust representation learning difficult and increasing interest in pretrained representations that can transfer useful visual structure beyond task-specific datasets.

Within this broader shift, foundation models emphasise large-scale pretraining and downstream adaptability~\cite{bommasani2021opportunities}. In general vision, Vision Transformers, masked autoencoders, and vision-language contrastive models have shown strong transferability at scale~\cite{dosovitskiy2020image,he2022masked,radford2021learning}. In medical imaging, resources and models such as RadImageNet, Models Genesis, and MedicalNet or Med3D seek to address label scarcity while preserving domain-specific and volumetric structure~\cite{mei2022radimagenet,zhou2019models,chen2019med3d}. More recent developments, including SAM, MedSAM, MedCLIP, and Merlin, further illustrate the rapid growth of large-scale pretrained medical representations~\cite{kirillov2023segment,ma2024segment,wang2022medclip,blankemeier2026merlin}. In this landscape, MedVAE is particularly relevant for RCC CT because it provides large-scale pretrained 2D and 3D variational autoencoders for medical image encoding~\cite{medvae}. Its emphasis on compact and generalisable latent representations makes it a plausible source of foundation representations for downstream classification when full-resolution learning is costly or data are limited.

Despite these advances, three gaps remain for CT-based RCC classification. First, it is unclear whether foundation representations alone are sufficiently discriminative in typical RCC cohort sizes, or whether radiomics still provides complementary information~\cite{aerts2014decoding,gillies2016radiomics,medvae}. Second, although RCC CT is inherently volumetric, many pipelines still operate on 2D slices, and the relative advantages of 2D slice-based versus 3D volumetric encoders, as well as their interaction with radiomics, remain insufficiently characterised~\cite{zhou2019models,chen2019med3d}. Third, although fusion may improve performance, it remains underexplored how to obtain clinically meaningful interpretability that reflects the internal decision pathway rather than relying only on post hoc visual explanations.

To examine radiomics--foundation interaction, we evaluate fusion as a comparative analytical tool. Specifically, we consider feature concatenation as a transparent baseline, cross-attention as a mechanism for modelling conditional dependence, and gated fusion as an established dynamic weighting strategy inspired by prior work on gated feature integration~\cite{vaswani2017attention,arevalo2017gated,baltruvsaitis2018multimodal}.

Interpretability is a further prerequisite for clinical adoption of medical imaging AI. Many explainability techniques are post hoc, including saliency maps~\cite{simonyan2013saliency}, Grad-CAM~\cite{selvaraju2017gradcam}, Integrated Gradients~\cite{sundararajan2017axiomatic}, LIME~\cite{ribeiro2016lime}, and SHAP~\cite{lundberg2017shap}. However, such explanations can remain visually plausible while being only weakly tied to learned parameters and may not faithfully reflect model reasoning~\cite{adebayo2018sanity}. In healthcare, multiple critiques and surveys have argued that current explainability methods can be misaligned with clinical needs and may offer limited patient-level decision support~\cite{ghassemi2021falsehope,vandervelden2022xai,borys2023xai}. In high-stakes settings, analysing internal decision processes, or using inherently interpretable approaches, has therefore been argued to be preferable to explaining black boxes after the fact~\cite{rudin2019stop,doshivelez2017rigorous}. In feature fusion settings, internal contribution analysis offers a practical route to decision-centric interpretability by quantifying how structured radiomics descriptors contribute within the fusion mechanism~\cite{baltruvsaitis2018multimodal,arevalo2017gated,zwanenburg2020image}.

In this work, we make the following contributions:

\begin{itemize}
    \item We present a CT-based RCC classification framework that combines MedVAE-derived 2D or 3D representations with handcrafted radiomics features under consistent preprocessing and evaluation.
    \item We compare 2D slice-based and 3D volumetric MedVAE encoders to assess how representation dimensionality affects RCC subtype discrimination.
    \item We evaluate representative fusion mechanisms, including concatenation, cross-attention, and gated fusion, to quantify radiomics--foundation interaction gains and stability.
    \item We assess the added value of radiomics beyond foundation representations and analyse radiomics feature importance within the fusion model to provide decision-centric interpretability.
\end{itemize}

\section{Methods}
\label{sec:methods}

\begin{figure}[!h]
  \centering
  \includegraphics[width=0.95\linewidth]{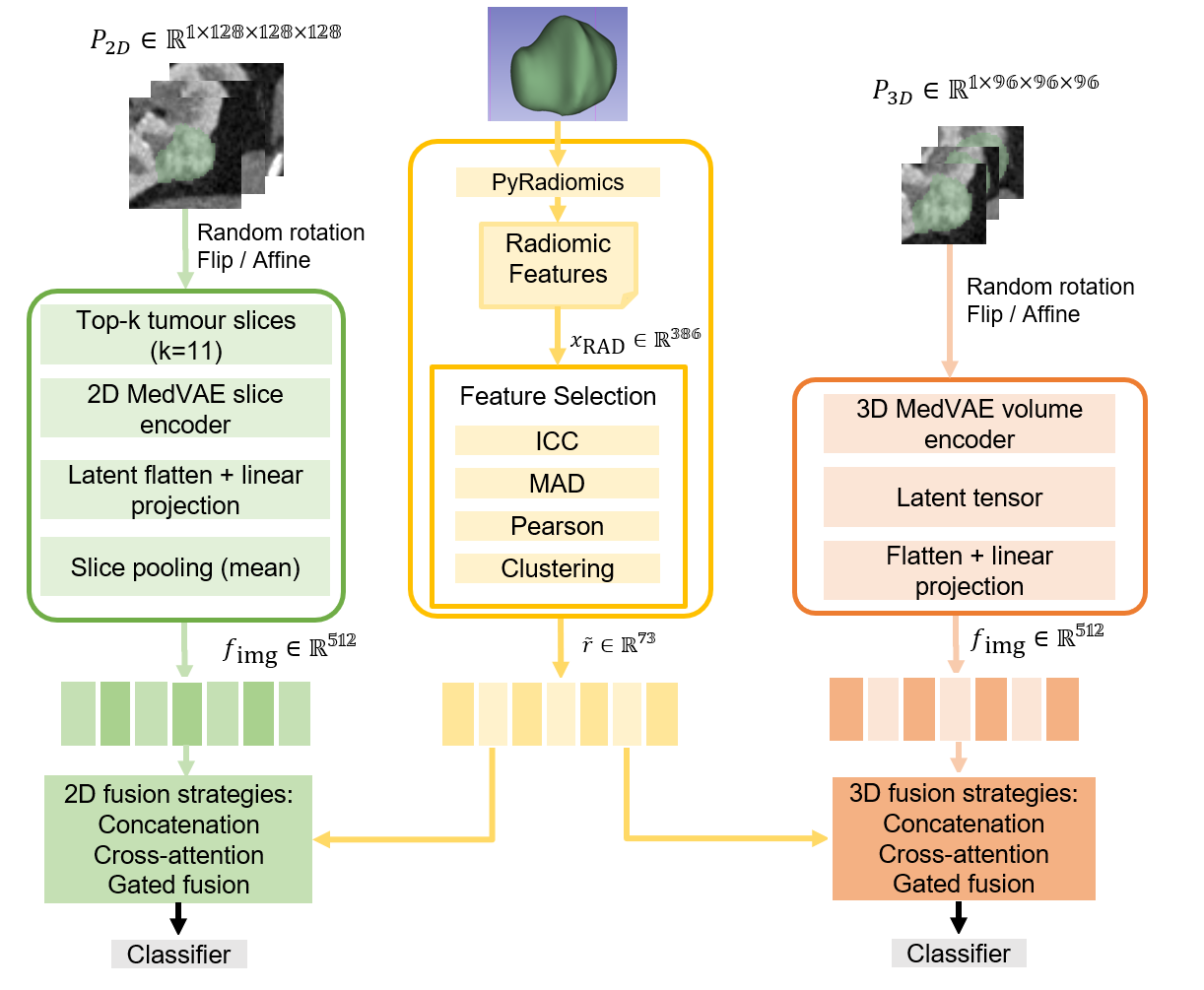}
  \caption{Overview of the radiomics--MedVAE fusion framework. Tumour-centred CT inputs are processed through either a 2D MedVAE branch with top-$k$ slice selection or a 3D MedVAE branch with volumetric encoding, while handcrafted radiomics features are extracted from the radiomics ROI and reduced through feature selection. The resulting image and radiomics representations are then combined using concatenation, cross-attention, or gated fusion for RCC subtype classification.}
  \label{fig:arch}
\end{figure}

\noindent\textbf{Methods overview.}
Figure~\ref{fig:arch} summarises the overall study framework. We compare handcrafted CT radiomics with MedVAE-derived foundation representations for RCC subtype classification on KiTS23. Under a unified preprocessing protocol, we extract (i) a structured radiomics vector from a tumour ROI, and (ii) a MedVAE image embedding using either a 2D slice-based pathway or a 3D volumetric pathway. We then evaluate three fusion strategies, concatenation, cross-attention, and gated fusion, and perform fusion-based interpretability analyses by ranking radiomics feature importance within the model decision pathway.

\subsection{Dataset and preprocessing}
\label{subsec:dataset_preprocessing}

We use the KiTS23 cohort, a publicly released challenge dataset of contrast-enhanced preoperative CT scans with semantic segmentations of kidney, tumour, and cyst structures~\cite{kits23website,heller2023kits23}. KiTS23 has a heterogeneous contrast setting, with cases acquired in either corticomedullary or nephrogenic phase~\cite{kits23website}. After excluding multifocal cases to avoid label ambiguity and intra-patient lesion heterogeneity, the final cohort used in this study comprised 396 cases.

All CT volumes and segmentation masks are reoriented to a common RAS axis convention and resampled to an isotropic voxel spacing of 1.0 mm using bilinear interpolation for images and nearest-neighbour interpolation for masks. CT intensities are clipped to a fixed Hounsfield Unit range of [-150, 200] prior to downstream processing.

For each case, we compute a tight tumour bounding box from the binary tumour mask. Based on this box, we generate two ROIs: (i) a \emph{radiomics ROI} using the tumour box without additional margin, and (ii) a \emph{deep-learning ROI} obtained by expanding the tumour box with a fixed 12-voxel margin to retain limited surrounding context. When the expanded ROI extends beyond the image boundary, padding is handled during subsequent patch construction.

Using the preprocessed case-level image volume and binary tumour mask, we generate a zero-margin segmentation crop for radiomics and an expanded crop for deep learning. If multiple tumour annotations are available for a case, we use a consensus tumour mask derived using Simultaneous Truth and Performance Level Estimation (STAPLE) together with the corresponding preprocessed case-level imaging volume to ensure geometric consistency~\cite{warfield2004staple}.

\subsection{Radiomics feature extraction and normalisation}
\label{subsec:radiomics}

As illustrated in Fig.~\ref{fig:arch}, radiomics descriptors are extracted from the \emph{radiomics ROI}, which is derived from the MONAI-preprocessed CT volumes and masks described above, using PyRadiomics~\cite{cardoso2022monai,van2017computational}. Feature computation follows IBSI-aligned settings to improve reproducibility~\cite{zwanenburg2020image}. We compute (i) shape features from the Original image type only, and (ii) first-order and texture features from both the Original image and Laplacian-of-Gaussian filtered images with $\sigma\in\{1,2,3\}$. This yields an initial radiomics feature vector $x_{\mathrm{RAD}}\in\mathbb{R}^{386}$.

Intensity discretisation is performed using a fixed bin width of 25 HU during radiomics extraction. To improve robustness, we adopt a multi-stage feature selection strategy. First, we retain only features with inter-observer reliability $\mathrm{ICC}\geq 0.75$. Second, we remove the lowest 20\% of features ranked by median absolute deviation. Third, highly correlated feature pairs are pruned using a Pearson correlation threshold of $|\rho|>0.95$. Finally, hierarchical clustering based on correlation distance is applied with a distance threshold of 0.2, and representative features are retained from each cluster. After this procedure, the selected radiomics vector is denoted by $\tilde r\in\mathbb{R}^{73}$. For fusion-based deep models, $\tilde r$ is z-score standardised using training-set statistics before projection and integration with MedVAE representations. By contrast, the standalone radiomics-based machine-learning baseline uses the selected radiomics features in their extracted form.

\subsection{MedVAE foundation representations}
\label{subsec:medvae}

As illustrated in Fig.~\ref{fig:arch}, we use pretrained MedVAE autoencoders as compact representation extractors for both the 2D and 3D image branches~\cite{medvae}. We compared frozen, partially fine-tuned, and fully fine-tuned encoder settings, and found partial fine-tuning to perform best. Unless otherwise stated, all reported MedVAE results therefore use the partially fine-tuned configuration, with selected pretrained modules updated jointly with the projection, fusion, and classification components on KiTS23.

Given an input $x$, the VAE encoder maps the image to a latent representation~\cite{kingma2014vae}. In practice, we use the latent tensor returned by the MedVAE forward pass. When the returned object provides a distribution-like mean, we use that mean as the image representation. The resulting latent tensor is then flattened and projected into a 512-dimensional shared embedding space.

\subsubsection{2D MedVAE pathway: slice selection and pooling}
\label{subsubsec:medvae2d}

For the 2D branch shown in Fig.~\ref{fig:arch}, we use the \emph{deep-learning ROI} after padding and centre-cropping to a fixed spatial size of $128\times128\times128$, yielding a single-case input tensor $P_{2D}\in\mathbb{R}^{1\times128\times128\times128}$. Axial slices are then ranked according to tumour extent in the mask, measured as the number of tumour-positive pixels in each slice. We retain the top-$k$ tumour-bearing slices with the largest tumour area and restore them to ascending axial order before encoding. Each selected slice is encoded by the pretrained 2D MedVAE encoder. The resulting slice-level latent tensor is flattened and projected to a 512-dimensional embedding,
\begin{equation}
u_i\in\mathbb{R}^{512}.
\end{equation}
Slice embeddings are then aggregated into a case-level image representation $f_{\mathrm{img}}\in\mathbb{R}^{512}$.

For the 2D branch, we adopt partial fine-tuning by freezing most pretrained MedVAE parameters and allowing only a subset of later pretrained modules to update during downstream training.

\subsubsection{3D MedVAE pathway: tumour-centred volumetric patch}
\label{subsubsec:medvae3d}

For the 3D branch shown in Fig.~\ref{fig:arch}, we use one tumour-centred volumetric crop per case. The expanded \emph{deep-learning ROI} is padded and centre-cropped to a fixed size of $96\times96\times96$, yielding an input tensor $P_{3D}\in\mathbb{R}^{1\times96\times96\times96}$. This patch is passed through the pretrained 3D MedVAE encoder, the returned latent tensor is flattened, and a learned linear projection maps it to a 512-dimensional image embedding $f_{\mathrm{img}}\in\mathbb{R}^{512}$. For the 3D branch, we adopt the same partial fine-tuning policy used in the 2D branch, freezing most pretrained MedVAE parameters and allowing only a subset of later pretrained modules to update during downstream training. The code also supports frozen and fully fine-tuned settings.

\subsection{Fusion models}
\label{subsec:fusion}

As illustrated in Fig.~\ref{fig:arch}, the selected radiomics feature vector $\tilde r\in\mathbb{R}^{73}$ and the MedVAE image embedding $f_{\mathrm{img}}\in\mathbb{R}^{512}$ are combined using three fusion strategies. Before fusion, radiomics features are projected into the same shared embedding space:
\begin{equation}
h_{\mathrm{rad}}=W_{\mathrm{rad}}\tilde r+b_{\mathrm{rad}},
\quad
W_{\mathrm{rad}}\in\mathbb{R}^{512\times 73}.
\label{eq:radproj}
\end{equation}

\subsubsection{Concatenation}
Concatenation directly combines the image embedding and projected radiomics embedding:
\begin{equation}
h_{\mathrm{fuse}}=[f_{\mathrm{img}};h_{\mathrm{rad}}],
\end{equation}
which is then passed to the classifier head for prediction.

\subsubsection{Cross-attention fusion}
The implemented cross-attention mechanism uses image token(s) as queries and radiomics tokens as keys and values. Radiomics features are first projected into a shared embedding and then transformed into a small set of learned radiomics tokens.

For the 2D branch, let $U\in\mathbb{R}^{k\times512}$ denote the slice-level image embeddings before pooling, and let $h_{\mathrm{rad}}\in\mathbb{R}^{512}$ denote the projected radiomics embedding. We form image queries and radiomics tokens as
\begin{align}
Q &= U W_Q, \\
R &= \mathrm{reshape}(h_{\mathrm{rad}} W_R),
\end{align}
where $Q\in\mathbb{R}^{k\times d_c}$ and $R\in\mathbb{R}^{T\times d_c}$. Cross-attention is then computed as
\begin{equation}
\mathrm{Attn}(Q,R,R)=\mathrm{softmax}\!\left(\frac{Q R^\top}{\sqrt{d_c}}\right)R.
\end{equation}
Residual connections, layer normalisation, and a feed-forward block are applied, after which the attended slice tokens are mean-pooled and concatenated with the pooled image embedding for classification.

For the 3D branch, the same principle is used, but the image side consists of a single pooled volumetric query token rather than multiple slice tokens. Let $q_{\mathrm{img}}\in\mathbb{R}^{1\times d_c}$ denote the projected volumetric query token and let $R\in\mathbb{R}^{T\times d_c}$ denote the radiomics tokens. Cross-attention is applied in the same form:
\begin{equation}
\mathrm{Attn}(q_{\mathrm{img}},R,R)=\mathrm{softmax}\!\left(\frac{q_{\mathrm{img}} R^\top}{\sqrt{d_c}}\right)R.
\end{equation}
The attended output is then fused with the image embedding for final classification.

\subsubsection{Gated fusion}
We implement channel-wise dynamic weighting inspired by prior work on gated feature integration~\cite{arevalo2017gated}. In the implemented formulation, the gate is generated from the radiomics projection alone:
\begin{equation}
g=\sigma\!\left(\mathrm{MLP}(h_{\mathrm{rad}})\right), \quad g\in(0,1)^{512},
\end{equation}
where a multilayer perceptron (MLP) uses a hidden layer of dimension 256 with ReLU activation, followed by a sigmoid output layer. The fused embedding is then
\begin{equation}
h_{\mathrm{fuse}} = g\odot f_{\mathrm{img}} + (1-g)\odot h_{\mathrm{rad}}.
\label{eq:gated}
\end{equation}
This formulation yields an explicit channel-wise balance between image-derived and radiomics-derived evidence.

\subsection{Fusion-based interpretability analysis}
\label{subsec:interpretability}

Post hoc saliency methods can be visually appealing but may be weakly coupled to the actual decision parameters of the trained model and can therefore be misleading in high-stakes settings~\cite{adebayo2018sanity,ghassemi2021falsehope,vandervelden2022xai,borys2023xai,rudin2019stop}. We therefore focus on decision-centric interpretability by analysing radiomics contributions inside the fusion model.

To quantify feature contribution, we compute permutation feature importance by randomly permuting one radiomics feature across cases in the evaluation split and measuring the resulting decrease in AUC. In the provided implementation, the permutation procedure is repeated $R=5$ times per feature, and the mean AUC drop is used for ranking.

\section{Experiments and Results}
\label{sec:experiments_results}

\subsection{Experimental setup}
\label{subsec:experimental_setup}

We optimise weighted cross-entropy loss for binary classification, where class weights are computed from the training split to mitigate label imbalance.

Both 2D and 3D implementations use Adam. In the 2D branch, all trainable parameters use a learning rate of $1\times10^{-4}$. In the 3D branch, the partially fine-tuned MedVAE backbone uses a learning rate of $1\times10^{-5}$, whereas newly added heads use $1\times10^{-4}$, with weight decay $1\times10^{-5}$. Training uses a \texttt{ReduceLROnPlateau} scheduler with factor 0.5 and patience 5, and early stopping based on validation AUC with patience 15.

Augmentations are applied only to the deep-learning branch. Training-time augmentation includes random flips, random 90-degree rotations, and small affine perturbations, whereas validation and test inputs are processed deterministically. We perform stratified case-level splitting to avoid information leakage, using fixed proportions of 60\% for training, 25\% for validation, and 15\% for testing.

The primary metric is the area under the receiver operating characteristic curve (AUC). We additionally report accuracy, F1-score, precision, recall (sensitivity), and specificity. Model selection is based on validation AUC. For threshold-dependent metrics, the final operating point is selected on the validation set using Youden's $J$ statistic and then applied unchanged to the held-out test set~\cite{youden1950index}.

\subsection{Overall performance comparison}
\label{subsec:overall_performance}

\begin{table}[h]
\centering
\caption{Performance comparison of radiomics, single-modality MedVAE, and fusion models for binary ccRCC versus non-ccRCC classification. The best result in each column is shown in bold. All metrics are reported in \%.}
\label{tab:main_results}
\begin{tabular}{llcccccc}
\hline
Category & Model & AUC & Accuracy & F1 & Precision & Recall & Specificity \\
\hline
Radiomics & ML baseline & 74.4 & 62.0 & 65.6 & \textbf{88.3} & 52.6 & \textbf{83.8} \\
\hline
\multirow{2}{*}{Deep learning}
& 2D MedVAE  & 59.5 & 63.3 & 75.0 & 71.7 & 78.6 & 27.8 \\
& 3D MedVAE & 63.2 & 71.7 & 82.5 & 72.7 & \textbf{95.2} & 16.7 \\
\hline
\multirow{3}{*}{2D fusion}
& Concatenation & 74.7 & 66.7 & 73.7 & 82.4 & 66.7 & 66.7 \\
& Cross-attention & 77.1 & 73.3 & 81.8 & 78.3 & 85.7 & 44.4 \\
& Gated fusion & 79.6 & \textbf{76.7} & \textbf{84.4} & 79.2 & 90.5 & 44.4 \\
\hline
\multirow{3}{*}{3D fusion}
& Concatenation & 73.8 & 61.7 & 67.6 & 82.8 & 57.1 & 72.2 \\
& Cross-attention & 79.5 & 71.7 & 79.5 & 80.5 & 78.6 & 55.6 \\
& Gated fusion & \textbf{82.7} & 71.7 & 77.9 & 85.7 & 71.4 & 72.2 \\
\hline
\end{tabular}
\end{table}

Table~\ref{tab:main_results} summarises the performance of radiomics, single-modality MedVAE, and fusion-based models for binary ccRCC versus non-ccRCC classification. Among all evaluated models, the 3D gated fusion model achieved the highest AUC of 82.7\%, indicating that joint modelling of volumetric MedVAE representations and handcrafted radiomics provided the strongest overall discrimination. The best 2D fusion model, also based on gated fusion, reached an AUC of 79.6\%, outperforming both the 2D-only model and the radiomics baseline.

Fusion improved performance consistently over single-modality image models. In the 2D setting, AUC increased from 59.5\% for the image-only branch to 74.7\%, 77.1\%, and 79.6\% for concatenation, cross-attention, and gated fusion, respectively. A similar pattern was observed for the 3D branch, where the image-only model achieved an AUC of 63.2\%, while the corresponding fusion variants reached 73.8\%, 79.5\%, and 82.7\%. These results suggest that radiomics provided complementary information that was not fully captured by MedVAE image representations alone.

Comparing fusion strategies, gated fusion yielded the strongest AUC in both the 2D and 3D settings. This trend suggests that adaptive channel-wise weighting offered a more effective integration mechanism than either simple concatenation or the current cross-attention formulation. Notably, the 3D gated fusion model also achieved competitive precision (85.7\%) and specificity (72.2\%), while maintaining balanced recall (71.4\%), indicating that its gain in AUC was not achieved through a highly skewed operating point.

The radiomics baseline remained competitive relative to single-modality image branches. The ML baseline achieved an AUC of 74.4\%, exceeding both the 2D-only and 3D-only MedVAE branches. This finding is consistent with the view that handcrafted radiomics remains a strong and stable descriptor family in relatively small RCC cohorts with heterogeneous acquisition conditions. However, the best fusion models outperformed the radiomics baseline, indicating that radiomics and MedVAE features were complementary rather than redundant.

\subsection{Ablation analysis}
\label{subsec:ablation}

\begin{table}[h]
\centering
\caption{Ablation analysis of the best-performing 3D gated fusion model. AUC is reported on the test set. The full fusion model is compared against image-only and radiomics-only variants using the same evaluation protocol.}
\label{tab:ablation_auc}
\begin{tabular}{lccc}
\hline
Variant & Image & Radiomics & AUC (\%) \\
\hline
Radiomics-only & $\times$ & $\checkmark$ & 52.9 \\
Image-only & $\checkmark$ & $\times$ & 60.4 \\
Full fusion & $\checkmark$ & $\checkmark$ & \textbf{82.7} \\
\hline
\end{tabular}
\end{table}

To further evaluate modality complementarity, we analysed the best-performing configuration using single-modality ablations. As shown in Table~\ref{tab:ablation_auc}, the full 3D gated fusion model achieved an AUC of 82.7\%, whereas the image-only and radiomics-only variants reached 60.4\% and 52.9\%, respectively. These values were obtained from the exported ablation results of the final model.

The large drop in AUC observed after removing either branch indicates that the performance of the final model cannot be explained by one modality alone. Instead, the gain from 60.4\% or 52.9\% to 82.7\% supports the interpretation that the learned interaction between radiomics and image representations was itself beneficial. In this setting, the image-only ablation outperformed the radiomics-only ablation, suggesting that the volumetric MedVAE branch provided the stronger standalone signal within the best 3D fusion configuration, while radiomics supplied additional structured information that substantially improved discrimination when fused.

\subsection{Fusion-based interpretability results}
\label{subsec:interpretability_results}

\begin{figure}[h]
  \centering
  \includegraphics[width=\linewidth]{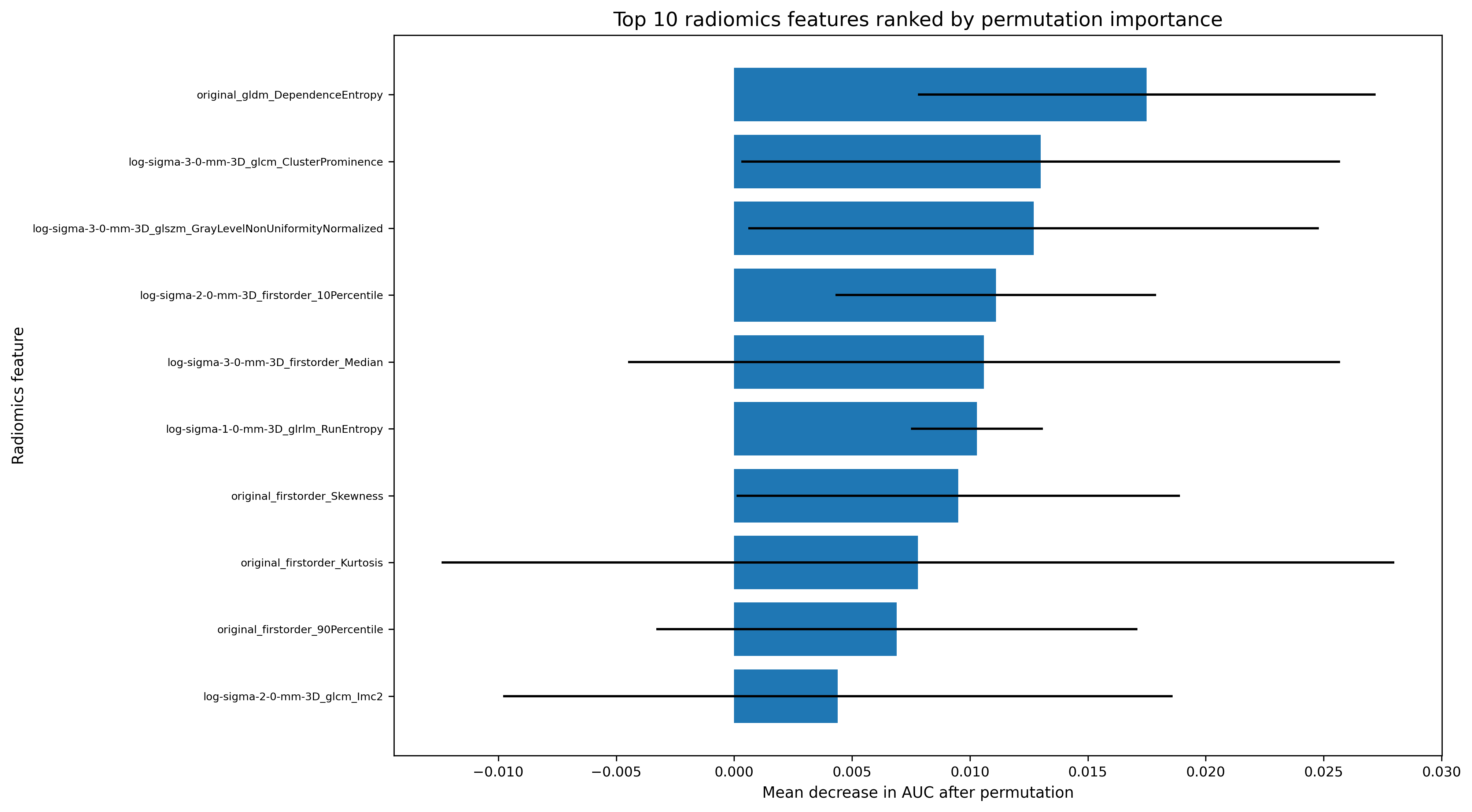}
  \caption{Top 10 radiomics features ranked by permutation importance for the best-performing 3D gated fusion model. Features are ordered by the mean decrease in AUC after feature-wise permutation on the evaluation set. Error bars indicate the standard deviation across repeated permutations.}
  \label{fig:feature_importance}
\end{figure}

To examine how radiomics contributed within the fusion model, we analysed permutation-based feature importance for the best-performing 3D gated fusion configuration. Figure~\ref{fig:feature_importance} presents the top 10 radiomics features ranked by the mean decrease in AUC after feature-wise permutation. The most influential feature was \texttt{original\_gldm\_DependenceEntropy}, while other highly ranked features came from GLCM, GLSZM, and first-order families, many of which were derived from Laplacian-of-Gaussian filtered images. These rankings were used to characterise which structured radiomics descriptors contributed most strongly within the fusion model.

Several qualitative patterns emerged from this ranking. First, many of the highest-ranked features belonged to texture families such as GLDM, GLCM, GLSZM, and GLRLM, suggesting that intratumoural heterogeneity remained a key discriminative factor in the fused model. Second, multiple top-ranked features were derived from Laplacian-of-Gaussian filtered images rather than from the original image alone, indicating that multi-scale filtered radiomics contributed useful information beyond simple first-order intensity summaries. Third, both texture and first-order statistics appeared among the most important features, suggesting that the final predictor relied on a combination of heterogeneity-related and distribution-related tumour descriptors.

This analysis supports the interpretability framing adopted in this study. Rather than relying exclusively on post hoc saliency visualisation, the fusion model permits direct examination of how structured radiomics descriptors participate in prediction. In this sense, feature-importance ranking provides a decision-centric view of the internal fusion mechanism, highlighting which radiomics characteristics most strongly influenced the final classifier output.

\section{Discussion}
\label{sec:discussion}

This study shows that radiomics remains valuable for RCC subtype classification even when MedVAE-derived foundation representations are available. Although the detailed performance comparisons are reported in the Results section, the overall pattern was consistent: integrating radiomics with pretrained image representations was more effective than relying on either branch alone. This supports the view that radiomics and foundation representations provide complementary rather than competing information in limited-data renal CT analysis.

The stronger performance of the 3D models further suggests that preserving volumetric structure is important for this task. RCC phenotype is inherently three-dimensional, and subtype-related heterogeneity is distributed across the tumour volume rather than confined to a small number of axial slices. In this setting, the 3D branch appears better aligned with the underlying anatomy, whereas the 2D branch necessarily compresses volumetric information through slice selection and pooling.

A particularly important aspect of this study is interpretability. Rather than relying only on post hoc saliency visualisation, we examined radiomics contribution within the fusion model itself. Because radiomics features correspond to defined intensity and texture descriptors, their ranking provides a more structured and clinically meaningful view of model behaviour~\cite{adebayo2018sanity,rudin2019stop,ghassemi2021falsehope,vandervelden2022xai,borys2023xai}. In practice, this helps connect model predictions to tumour characteristics that are more interpretable than abstract latent features alone. Such a decision-centric perspective may be particularly valuable in preoperative decision support, where clinicians need not only a prediction, but also some indication of which tumour properties are influencing it.

This interpretability advantage may also be relevant beyond RCC, as similar fusion designs could support other medical imaging tasks that require both strong discrimination and clinically communicable model behaviour.

Overall, these findings support a complementary role for radiomics in the foundation-model era. For RCC CT classification, the best results were obtained when MedVAE-derived image features were fused with radiomics, especially in the 3D setting. More broadly, fusion with structured radiomics may offer a practical route toward clinically useful medical AI that balances predictive performance with more interpretable decision support.
\section{Conclusion}
\label{sec:conclusion}

In this study, we investigated the interaction between handcrafted radiomics and MedVAE-derived foundation representations for RCC subtype classification from contrast-enhanced CT. The results show that radiomics remains clinically relevant even in the presence of pretrained image representations, and that the strongest performance is achieved when the two are fused rather than used in isolation. The best overall model was obtained in the 3D setting, supporting the value of preserving volumetric tumour structure for RCC characterisation.

Beyond predictive performance, the proposed framework also provides a more structured route to interpretability by identifying which radiomics descriptors contribute within the fusion model. This is clinically important because preoperative subtype classification requires not only accurate discrimination, but also outputs that can be related to recognisable tumour characteristics. More broadly, the same fusion paradigm may be transferable to other medical imaging tasks in which pretrained image representations can be complemented by structured radiomics to improve both performance and interpretability.

\bibliographystyle{unsrt}
\bibliography{references}

@unknown{medvae,
author = {Varma, Maya and Kumar, Ashwin and Sluijs, Rogier and Ostmeier, Sophie and Blankemeier, Louis and Chambon, Pierre and Blüthgen, Christian and Prince, Jip and Langlotz, Curtis and Chaudhari, Akshay},
year = {2025},
month = {02},
pages = {},
title = {MedVAE: Efficient Automated Interpretation of Medical Images with Large-Scale Generalizable Autoencoders},
doi = {10.48550/arXiv.2502.14753}
}

@inproceedings{kirillov2023segment,
  title={Segment anything},
  author={Kirillov, Alexander and Mintun, Eric and Ravi, Nikhila and Mao, Hanzi and Rolland, Chloe and Gustafson, Laura and Xiao, Tete and Whitehead, Spencer and Berg, Alexander C and Lo, Wan-Yen and others},
  booktitle={Proceedings of the IEEE/CVF international conference on computer vision},
  pages={4015--4026},
  year={2023}
}

@article{ma2024segment,
  title={Segment anything in medical images},
  author={Ma, Jun and He, Yuting and Li, Feifei and Han, Lin and You, Chenyu and Wang, Bo},
  journal={Nature communications},
  volume={15},
  number={1},
  pages={654},
  year={2024},
  publisher={Nature Publishing Group UK London}
}

@article{mei2022radimagenet,
  title={RadImageNet: an open radiologic deep learning research dataset for effective transfer learning},
  author={Mei, Xueyan and Liu, Zelong and Robson, Philip M and Marinelli, Brett and Huang, Mingqian and Doshi, Amish and Jacobi, Adam and Cao, Chendi and Link, Katherine E and Yang, Thomas and others},
  journal={Radiology: Artificial Intelligence},
  volume={4},
  number={5},
  pages={e210315},
  year={2022},
  publisher={Radiological Society of North America}
}

@inproceedings{zhou2019models,
  title={Models genesis: Generic autodidactic models for 3d medical image analysis},
  author={Zhou, Zongwei and Sodha, Vatsal and Rahman Siddiquee, Md Mahfuzur and Feng, Ruibin and Tajbakhsh, Nima and Gotway, Michael B and Liang, Jianming},
  booktitle={International conference on medical image computing and computer-assisted intervention},
  pages={384--393},
  year={2019},
  organization={Springer}
}

@article{chen2019med3d,
  title={Med3d: Transfer learning for 3d medical image analysis},
  author={Chen, Sihong and Ma, Kai and Zheng, Yefeng},
  journal={arXiv preprint arXiv:1904.00625},
  year={2019}
}

@inproceedings{wang2022medclip,
  title={Medclip: Contrastive learning from unpaired medical images and text},
  author={Wang, Zifeng and Wu, Zhenbang and Agarwal, Dinesh and Sun, Jimeng},
  booktitle={Proceedings of the 2022 Conference on Empirical Methods in Natural Language Processing},
  pages={3876--3887},
  year={2022}
}

@article{blankemeier2026merlin,
  title={Merlin: a computed tomography vision--language foundation model and dataset},
  author={Blankemeier, Louis and Kumar, Ashwin and Cohen, Joseph Paul and Liu, Jiaming and Liu, Longchao and Van Veen, Dave and Gardezi, Syed Jamal Safdar and Yu, Hongkun and Paschali, Magdalini and Chen, Zhihong and others},
  journal={Nature},
  pages={1--11},
  year={2026},
  publisher={Nature Publishing Group UK London}
}

@article{bommasani2021opportunities,
  title={On the opportunities and risks of foundation models},
  author={Bommasani, Rishi and Hudson, Drew A and Adeli, Ehsan and Altman, Russ and Arora, Simran and von Arx, Sydney and Bernstein, Michael S and Bohg, Jeannette and Bosselut, Antoine and Brunskill, Emma and others},
  journal={arXiv preprint arXiv:2108.07258},
  year={2021}
}

@article{dosovitskiy2020image,
  title={An image is worth 16x16 words: Transformers for image recognition at scale},
  author={Dosovitskiy, Alexey and Beyer, Lucas and Kolesnikov, Alexander and Weissenborn, Dirk and Zhai, Xiaohua and Unterthiner, Thomas and Dehghani, Mostafa and Minderer, Matthias and Heigold, Georg and Gelly, Sylvain and others},
  journal={arXiv preprint arXiv:2010.11929},
  year={2020}
}

@inproceedings{he2022masked,
  title={Masked autoencoders are scalable vision learners},
  author={He, Kaiming and Chen, Xinlei and Xie, Saining and Li, Yanghao and Doll{\'a}r, Piotr and Girshick, Ross},
  booktitle={Proceedings of the IEEE/CVF conference on computer vision and pattern recognition},
  pages={16000--16009},
  year={2022}
}

@inproceedings{radford2021learning,
  title={Learning transferable visual models from natural language supervision},
  author={Radford, Alec and Kim, Jong Wook and Hallacy, Chris and Ramesh, Aditya and Goh, Gabriel and Agarwal, Sandhini and Sastry, Girish and Askell, Amanda and Mishkin, Pamela and Clark, Jack and others},
  booktitle={International conference on machine learning},
  pages={8748--8763},
  year={2021},
  organization={PmLR}
}

@article{litjens2017survey,
  title={A survey on deep learning in medical image analysis},
  author={Litjens, Geert and Kooi, Thijs and Bejnordi, Babak Ehteshami and Setio, Arnaud Arindra Adiyoso and Ciompi, Francesco and Ghafoorian, Mohsen and Van Der Laak, Jeroen Awm and Van Ginneken, Bram and S{\'a}nchez, Clara I},
  journal={Medical image analysis},
  volume={42},
  pages={60--88},
  year={2017},
  publisher={Elsevier}
}

@article{aerts2014decoding,
  title={Decoding tumour phenotype by noninvasive imaging using a quantitative radiomics approach},
  author={Aerts, Hugo JWL and Velazquez, Emmanuel Rios and Leijenaar, Ralph TH and Parmar, Chintan and Grossmann, Patrick and Carvalho, Sara and Bussink, Johan and Monshouwer, Ren{\'e} and Haibe-Kains, Benjamin and Rietveld, Derek and others},
  journal={Nature communications},
  volume={5},
  number={1},
  pages={4006},
  year={2014},
  publisher={Nature Publishing Group UK London}
}

@article{gillies2016radiomics,
  title={Radiomics: images are more than pictures, they are data},
  author={Gillies, Robert J and Kinahan, Paul E and Hricak, Hedvig},
  journal={Radiology},
  volume={278},
  number={2},
  pages={563--577},
  year={2016},
  publisher={Radiological Society of North America}
}

@article{lambin2017radiomics,
  title={Radiomics: the bridge between medical imaging and personalized medicine},
  author={Lambin, Philippe and Leijenaar, Ralph TH and Deist, Timo M and Peerlings, Jurgen and De Jong, Evelyn EC and Van Timmeren, Janita and Sanduleanu, Sebastian and Larue, Ruben THM and Even, Aniek JG and Jochems, Arthur and others},
  journal={Nature reviews Clinical oncology},
  volume={14},
  number={12},
  pages={749--762},
  year={2017},
  publisher={Nature Publishing Group UK London}
}

@article{zwanenburg2020image,
  title={The image biomarker standardization initiative: standardized quantitative radiomics for high-throughput image-based phenotyping},
  author={Zwanenburg, Alex and Valli{\`e}res, Martin and Abdalah, Mahmoud A and Aerts, Hugo JWL and Andrearczyk, Vincent and Apte, Aditya and Ashrafinia, Saeed and Bakas, Spyridon and Beukinga, Roelof J and Boellaard, Ronald and others},
  journal={Radiology},
  volume={295},
  number={2},
  pages={328--338},
  year={2020},
  publisher={Radiological Society of North America}
}

@article{van2017computational,
  title={Computational radiomics system to decode the radiographic phenotype},
  author={Van Griethuysen, Joost JM and Fedorov, Andriy and Parmar, Chintan and Hosny, Ahmed and Aucoin, Nicole and Narayan, Vivek and Beets-Tan, Regina GH and Fillion-Robin, Jean-Christophe and Pieper, Steve and Aerts, Hugo JWL},
  journal={Cancer research},
  volume={77},
  number={21},
  pages={e104--e107},
  year={2017},
  publisher={American Association for Cancer Research}
}

@article{wang2021radiomics,
  title={Radiomics models based on enhanced computed tomography to distinguish clear cell from non-clear cell renal cell carcinomas},
  author={Wang, Ping and Pei, Xu and Yin, Xiao-Ping and Ren, Jia-Liang and Wang, Yun and Ma, Lu-Yao and Du, Xiao-Guang and Gao, Bu-Lang},
  journal={Scientific Reports},
  volume={11},
  number={1},
  pages={13729},
  year={2021},
  publisher={Nature Publishing Group UK London}
}

@article{uhm2021deep,
  title={Deep learning for end-to-end kidney cancer diagnosis on multi-phase abdominal computed tomography},
  author={Uhm, Kwang-Hyun and Jung, Seung-Won and Choi, Moon Hyung and Shin, Hong-Kyu and Yoo, Jae-Ik and Oh, Se Won and Kim, Jee Young and Kim, Hyun Gi and Lee, Young Joon and Youn, Seo Yeon and others},
  journal={NPJ precision oncology},
  volume={5},
  number={1},
  pages={54},
  year={2021},
  publisher={Nature Publishing Group UK London}
}

@article{arevalo2017gated,
  title={Gated multimodal units for information fusion},
  author={Arevalo, John and Solorio, Thamar and Montes-y-G{\'o}mez, Manuel and Gonz{\'a}lez, Fabio A},
  journal={arXiv preprint arXiv:1702.01992},
  year={2017}
}

@article{baltruvsaitis2018multimodal,
  title={Multimodal machine learning: A survey and taxonomy},
  author={Baltru{\v{s}}aitis, Tadas and Ahuja, Chaitanya and Morency, Louis-Philippe},
  journal={IEEE transactions on pattern analysis and machine intelligence},
  volume={41},
  number={2},
  pages={423--443},
  year={2018},
  publisher={IEEE}
}

@misc{iarc2024kidneyfactsheet,
  author       = {{International Agency for Research on Cancer}},
  title        = {Kidney Fact Sheet},
  year         = {2024},
  howpublished = {\url{https://gco.iarc.who.int/media/globocan/factsheets/cancers/29-kidney-fact-sheet.pdf}},
  note         = {Global Cancer Observatory, GLOBOCAN 2022, version 1.1; accessed 13 April 2026}
}

@article{ljungberg2023eau,
  title={EAU guidelines on renal cell carcinoma},
  author={Ljungberg, B and Albiges, Laurence and Bedke, J and Bex, A and Capitanio, U and Giles, RH and Hora, M and Klatte, T and Marconi, L and Powles, T and others},
  journal={European Association of Urology},
  pages={1--100},
  year={2023}
}

@inproceedings{selvaraju2017gradcam,
  title     = {Grad-CAM: Visual Explanations from Deep Networks via Gradient-Based Localization},
  author    = {Selvaraju, Ramprasaath R. and Cogswell, Michael and Das, Abhishek and Vedantam, Ramakrishna and Parikh, Devi and Batra, Dhruv},
  booktitle = {Proceedings of the IEEE International Conference on Computer Vision (ICCV)},
  pages     = {618--626},
  year      = {2017},
  doi       = {10.1109/ICCV.2017.74}
}

@article{simonyan2013saliency,
  title   = {Deep Inside Convolutional Networks: Visualising Image Classification Models and Saliency Maps},
  author  = {Simonyan, Karen and Vedaldi, Andrea and Zisserman, Andrew},
  journal = {arXiv preprint arXiv:1312.6034},
  year    = {2013},
  url     = {https://arxiv.org/abs/1312.6034}
}

@inproceedings{sundararajan2017axiomatic,
  title     = {Axiomatic Attribution for Deep Networks},
  author    = {Sundararajan, Mukund and Taly, Ankur and Yan, Qiqi},
  booktitle = {Proceedings of the 34th International Conference on Machine Learning (ICML)},
  series    = {Proceedings of Machine Learning Research},
  volume    = {70},
  pages     = {3319--3328},
  year      = {2017},
  url       = {https://proceedings.mlr.press/v70/sundararajan17a.html}
}

@inproceedings{ribeiro2016lime,
  title     = {"Why Should I Trust You?": Explaining the Predictions of Any Classifier},
  author    = {Ribeiro, Marco Tulio and Singh, Sameer and Guestrin, Carlos},
  booktitle = {Proceedings of the 22nd ACM SIGKDD International Conference on Knowledge Discovery and Data Mining (KDD)},
  pages     = {1135--1144},
  year      = {2016},
  doi       = {10.1145/2939672.2939778}
}

@inproceedings{lundberg2017shap,
  title     = {A Unified Approach to Interpreting Model Predictions},
  author    = {Lundberg, Scott M. and Lee, Su-In},
  booktitle = {Advances in Neural Information Processing Systems},
  volume    = {30},
  year      = {2017},
  url       = {https://arxiv.org/abs/1705.07874}
}

@article{ghassemi2021falsehope,
  title   = {The false hope of current approaches to explainable artificial intelligence in health care},
  author  = {Ghassemi, Marzyeh and Oakden-Rayner, Luke and Beam, Andrew L.},
  journal = {The Lancet Digital Health},
  volume  = {3},
  number  = {11},
  pages   = {e745--e750},
  year    = {2021},
  doi     = {10.1016/S2589-7500(21)00208-9}
}

@article{vandervelden2022xai,
  title   = {Explainable artificial intelligence (XAI) in deep learning-based medical image analysis},
  author  = {van der Velden, Bas H. M. and Kuijf, Hugo J. and Gilhuijs, Kenneth G. A. and Viergever, Max A.},
  journal = {Medical Image Analysis},
  volume  = {79},
  pages   = {102470},
  year    = {2022},
  doi     = {10.1016/j.media.2022.102470}
}

@article{borys2023xai,
  title   = {Explainable AI in medical imaging: An overview for clinical practitioners -- Beyond saliency-based XAI approaches},
  author  = {Borys, Katarzyna and Schmitt, Yasmin Alyssa and Nauta, Meike and Seifert, Christin and Kr{\"a}mer, Nicole and Friedrich, Christoph M. and Nensa, Felix},
  journal = {European Journal of Radiology},
  volume  = {162},
  pages   = {110786},
  year    = {2023},
  doi     = {10.1016/j.ejrad.2023.110786}
}

@article{rudin2019stop,
  title   = {Stop explaining black box machine learning models for high stakes decisions and use interpretable models instead},
  author  = {Rudin, Cynthia},
  journal = {Nature Machine Intelligence},
  volume  = {1},
  pages   = {206--215},
  year    = {2019},
  doi     = {10.1038/s42256-019-0048-x}
}

@article{doshivelez2017rigorous,
  title   = {Towards A Rigorous Science of Interpretable Machine Learning},
  author  = {Doshi-Velez, Finale and Kim, Been},
  journal = {arXiv preprint arXiv:1702.08608},
  year    = {2017},
  url     = {https://arxiv.org/abs/1702.08608}
}

@misc{kits23website,
  title        = {KiTS23: The 2023 Kidney and Kidney Tumor Segmentation Challenge},
  author       = {{KiTS Challenge Organizers}},
  year         = {2023},
  howpublished = {\url{https://kits-challenge.org/kits23/}},
  note         = {Accessed: 2026-04-14}
}

@misc{heller2023kits23,
  title        = {2023 Kidney and Kidney Tumor Segmentation Challenge},
  author       = {Heller, Nicholas and Isensee, Fabian and Tejpaul, Resha and Wood, Andrew and Papanikolopoulos, Nikolaos and Weight, Christopher},
  year         = {2023},
  publisher    = {Zenodo},
  doi          = {10.5281/zenodo.7840134},
  url          = {https://doi.org/10.5281/zenodo.7840134}
}

@inproceedings{kingma2014vae,
  title     = {Auto-Encoding Variational Bayes},
  author    = {Kingma, Diederik P. and Welling, Max},
  booktitle = {International Conference on Learning Representations (ICLR)},
  year      = {2014},
  url       = {https://arxiv.org/abs/1312.6114}
}

@article{warfield2004staple,
  title   = {Simultaneous truth and performance level estimation (STAPLE): an algorithm for the validation of image segmentation},
  author  = {Warfield, Simon K. and Zou, Kelly H. and Wells, William M.},
  journal = {IEEE Transactions on Medical Imaging},
  volume  = {23},
  number  = {7},
  pages   = {903--921},
  year    = {2004},
  doi     = {10.1109/TMI.2004.828354}
}

@inproceedings{vaswani2017attention,
  title     = {Attention Is All You Need},
  author    = {Vaswani, Ashish and Shazeer, Noam and Parmar, Niki and Uszkoreit, Jakob and Jones, Llion and Gomez, Aidan N. and Kaiser, {\L}ukasz and Polosukhin, Illia},
  booktitle = {Advances in Neural Information Processing Systems (NeurIPS)},
  year      = {2017},
  url       = {https://papers.nips.cc/paper/7181-attention-is-all-you-need}
}

@inproceedings{adebayo2018sanity,
  title     = {Sanity Checks for Saliency Maps},
  author    = {Adebayo, Julius and Gilmer, Justin and Muelly, Michael and Goodfellow, Ian J. and Hardt, Moritz and Kim, Been},
  booktitle = {Advances in Neural Information Processing Systems (NeurIPS)},
  pages     = {9525--9536},
  year      = {2018},
  url       = {https://papers.nips.cc/paper/8160-sanity-checks-for-saliency-maps}
}

@article{cardoso2022monai,
  title={MONAI: An open-source framework for deep learning in healthcare},
  author={Cardoso, M. Jorge and Li, Wenqi and Brown, Richard and others},
  journal={IEEE Journal of Biomedical and Health Informatics},
  volume={26},
  number={9},
  pages={4123--4135},
  year={2022}
}

@article{youden1950index,
  title={Index for rating diagnostic tests},
  author={Youden, W. J.},
  journal={Cancer},
  volume={3},
  number={1},
  pages={32--35},
  year={1950},
  doi={10.1002/1097-0142(1950)3:1<32::AID-CNCR2820030106>3.0.CO;2-3}
}
\end{document}